\pdfoutput=1
\documentclass[11pt]{article}
\usepackage{authblk}
\usepackage{EMNLP2023}
\usepackage{times}
\usepackage{latexsym}
\usepackage[T1]{fontenc}
\usepackage[utf8]{inputenc}
\usepackage{microtype}
\usepackage{inconsolata}
\usepackage{tabularray}
\usepackage{tikz}

\usepackage{tcolorbox}
\usepackage{adjustbox}
\usepackage{amsmath}
\usepackage{amssymb}
\usepackage{graphicx}
\usepackage{caption}
\usepackage{subcaption}
\usepackage{multirow}
\usepackage{xcolor}
\usepackage{xcolor,amsmath}
\usepackage[linesnumbered,ruled,vlined]{algorithm2e}
\DontPrintSemicolon

\definecolor{color1}{HTML}{003F5C}
\definecolor{color2}{HTML}{374C80}
\definecolor{color3}{HTML}{00B7EB}
\definecolor{color4}{HTML}{BC5090}
\definecolor{color5}{HTML}{EF5675}
\definecolor{color6}{HTML}{FF764A}
\definecolor{color7}{HTML}{FFA600}

\UseTblrLibrary{booktabs}

\usetikzlibrary{positioning, arrows.meta, calc, shapes, fit}

\newcommand{\entity}[2]{{\textcolor{color1}{[}{#1}\textcolor{color1}{]}\textcolor{color1}{\small #2}}}

\title{Calibrated Seq2seq Models for Efficient and \\ Generalizable Ultra-fine Entity Typing}

\author{
 \textbf{Yanlin Feng}\textsuperscript{~$\dagger$~$\S$} \thanks{~~This work was done while the first author was at Carnegie
Mellon University.}~~
 \textbf{Adithya Pratapa}\textsuperscript{$\dagger$}~~
 \textbf{David Mortensen}\textsuperscript{$\dagger$} \\
 \textsuperscript{$\dagger$}{Language Technologies Institute, Carnegie Mellon University} \\ 
 \textsuperscript{$\S$}{Megagon Labs}  \\
 \texttt{yanlin@megagon.ai, \{vpratapa, dmortens\}@cs.cmu.edu} \\ 
}

\begin{document}
\maketitle
\begin{abstract}

   Ultra-fine entity typing plays a crucial role in information extraction by predicting fine-grained semantic types for entity mentions in text. However, this task poses significant challenges due to the massive number of entity types in the output space. The current state-of-the-art approaches, based on standard multi-label classifiers or cross-encoder models, suffer from poor generalization performance or inefficient inference. In this paper, we present CASENT, a seq2seq model designed for ultra-fine entity typing that predicts ultra-fine types with calibrated confidence scores. Our model takes an entity mention as input and employs constrained beam search to generate multiple types autoregressively. The raw sequence probabilities associated with the predicted types are then transformed into confidence scores using a novel calibration method. We conduct extensive experiments on the UFET dataset which contains over $10k$ types. Our method outperforms the previous state-of-the-art in terms of F1 score and calibration error, while achieving an inference speedup of over $50$ times. Additionally, we demonstrate the generalization capabilities of our model by evaluating it in zero-shot and few-shot settings on five specialized domain entity typing datasets that are unseen during training. Remarkably, our model outperforms large language models with 10 times more parameters in the zero-shot setting, and when fine-tuned on 50 examples, it significantly outperforms ChatGPT on all datasets.\footnote{Our code, models and demo are available at \url{https://github.com/yanlinf/CASENT}.}

\end{abstract}

\section{Introduction}

% \drm{Provide a line or to in the intro about your technical approach.}

% \ap{I wonder if we should say GPT-3.5 instead of ChatGPT. I feel ChatGPT is an umbrella term that includes GPT-4.}
% \ap{Should we use Seq2Seq instead of Seq2seq in the title?}
\begin{figure}[t]
    \centering
    \includegraphics[width=\linewidth]{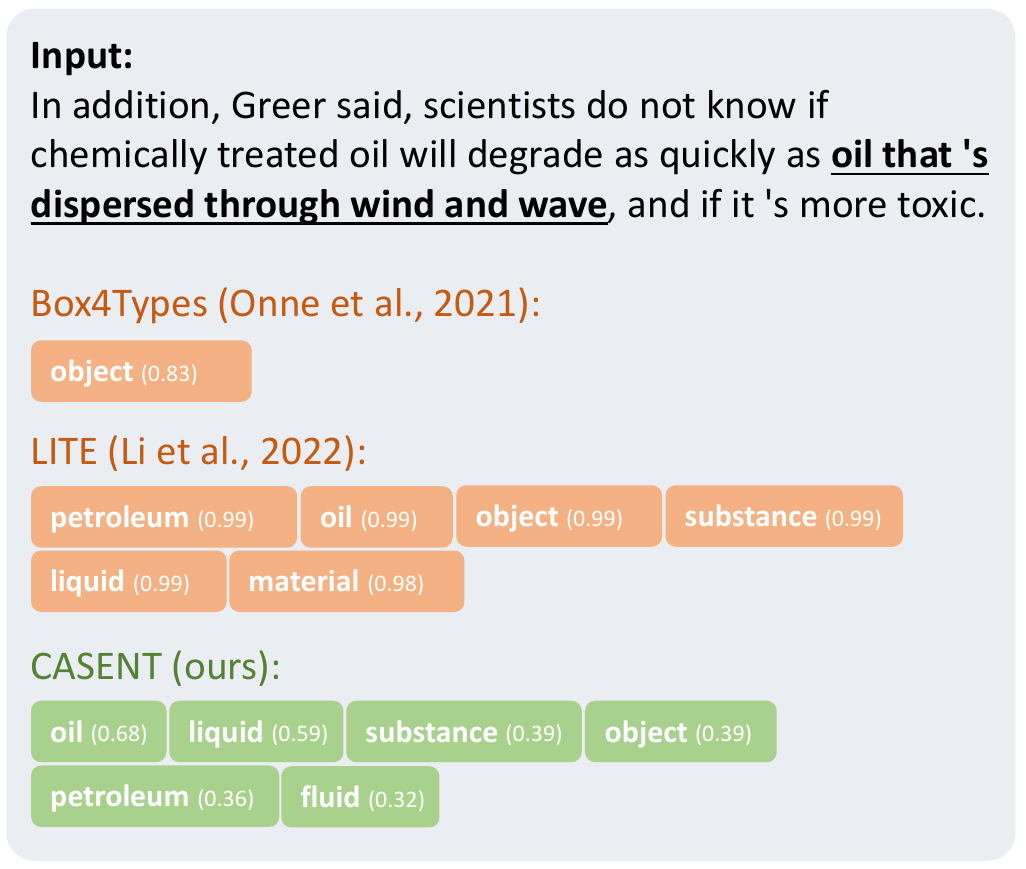}
    \caption{Comparison of predicted labels and confidence scores for a UFET test example using Box4Types \cite{onoe2021modeling}, LITE \cite{li2022ultra}, and our approach, CASENT. Predictions are sorted in descending order based on confidence. Box4Types fails to generalize to rare and unseen types, while LITE does not predict calibrated confidence scores and exhibits slow inference speed.}
    \label{fig:example}
\end{figure}

Classifying entities mentioned in text into types, commonly known as entity typing, is a fundamental problem in information extraction. Earlier research on entity typing focused on relatively small type inventories \cite{ling2012fine} which imposed severe limitations on the practical value of such systems, given the vast number of types in the real world. For example, WikiData, the current largest knowledge base in the world, records more than $2.7$ million entity types\footnote{Estimated from the unique children in the \textit{subclassOf (P279)} relations using the February 2023 Wikidata dump.}. As a result, a fully supervised approach will always be hampered by insufficient training data. Recently, \citet{choi2018ultra} introduced the task of ultra-fine entity typing (UFET), a multi-label entity classification task with over $10k$ fine-grained types. In this work, we make the first step towards building an efficient general-purpose entity typing model by leveraging the UFET dataset. Our model not only achieves state-of-the-art performance on UFET but also generalizes outside of the UFET type vocabulary. An example prediction of our model is shown in \autoref{fig:example}. 

% Ultra-fine entity typing, a multi-label entity classification task with over 10k fine-grained types, presents significant challenges due to the massive number of labels in the output space.}

% \sout{Entity typing systems are useful in various natural language processing (NLP) tasks such as entity linking \cite{onoe2020fine}, relation extraction \cite{yaghoobzadeh-etal-2017-noise}, coreference resolution \cite{onoe-durrett-2020-interpretable}, and question answering \cite{yavuz2016improving}. Earlier research on entity typing focused on relatively small type inventories \cite{ling2012fine} and only focused on named entities. Recently, \citet{choi2018ultra} proposed an ultra-fine entity typing (UFET) task that incorporates arbitrary noun phrases with an inventory of over $10k$ fine-grained semantic types. In this work, we present an efficient methodology for ultra-fine entity typing---for identifying the types of entities in text from a large (potentially open) set.}

Ultra-fine entity typing can be viewed as a multi-label classification problem over an extensive label space. A standard approach to this task employs multi-label classifiers that map contextual representations of the input entity mention to scores using a linear transformation \cite{choi2018ultra, dai2021ultra, onoe2021modeling}. While this approach offers superior inference speeds, it ignores the type semantics by treating all types as integer indices and thus fails to generalize to unseen types. The current state-of-the-art approach \cite{li2022ultra} reformulated entity typing as a textual entailment task. They presented a cross-encoder model that computes an entailment score between the entity mention and a candidate type. Despite its strong generalization capabilities, this approach is inefficient given the need to enumerate all $10k$ types in the UFET dataset.

Black-box large language models, such as GPT-3 and ChatGPT, have demonstrated impressive zero-shot and few-shot capabilities in a wide range of generation and understanding tasks \cite{brown2020language, ouyang2022training}. Yet, applying them to ultra-fine entity typing poses challenges due to the extensive label space and the context length limit of these models. For instance, \citet{zhan2023glen} reported that GPT-3 with few-shot prompting does not perform well on a classification task with thousands of classes. Similar observations have been made in our experiments conducted on UFET.
% \ap{might be a good idea to state here if we see similar behavior for UFET.}

\begin{figure*}[t]

    \centering
    \includegraphics[width=\textwidth]{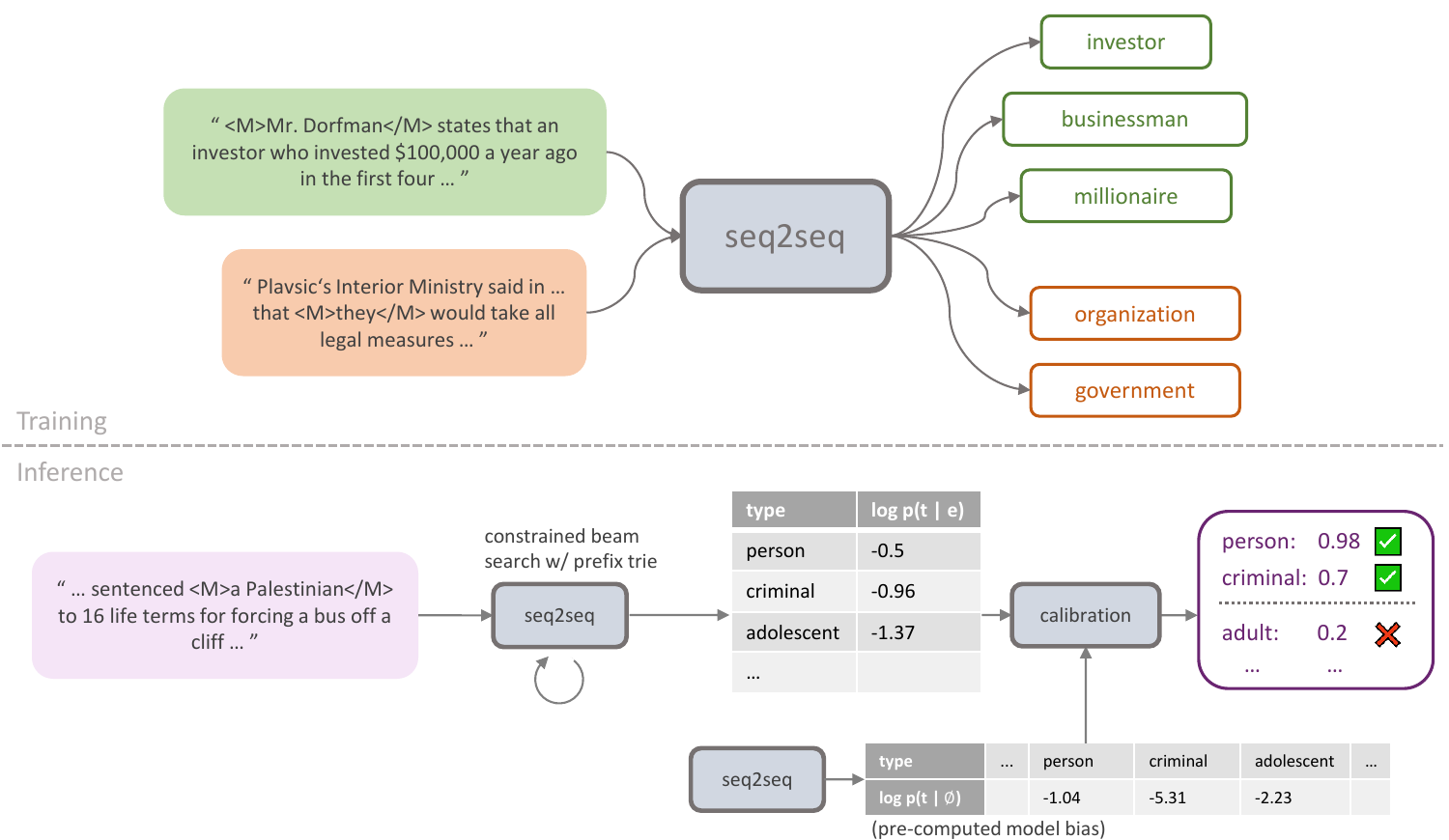}
    \caption{\small  Overview of the training and inference process of CASENT. We present an example output from our model.}
    \label{fig:model}
\end{figure*}

In this work, we propose CASENT, a \textbf{Ca}librated \textbf{S}eq2Seq model for \textbf{En}tity \textbf{T}yping. CASENT predicts ultra-fine entity types with calibrated confidence scores using a seq2seq model (T5-large \cite{raffel2020exploring}). Our approach offers several advantages compared to previous methods: (1) Standard maximum likelihood training without the need for negative sampling or sophisticated loss functions (2) Efficient inference through a single autoregressive decoding pass (3) Calibrated confidence scores that align with the expected accuracy of the predictions (4) Strong generalization performance to unseen domains and types. An illustration of our approach is provided in \autoref{fig:model}.

While seq2seq formulation has been successfully applied to NLP tasks such as entity linking \cite{de2020autoregressive, de2022multilingual}, its application to ultra-fine entity typing remains non-trivial due to the multi-label prediction requirement. A simple adaptation would employ beam search to decode multiple types and use a probability threshold to select types. However, we show that this approach fails to achieve optimal performance as the raw conditional probabilities do not align with the true likelihood of the corresponding types. In this work, we propose to transform the raw probabilities into calibrated confidence scores that reflect the true likelihood of the decoded types. To this end, we extend Platt scaling \cite{platt1999probabilistic}, a standard technique for calibrating binary classifiers, to the multi-label setting. To mitigate the label sparsity issue in ultra-fine entity typing, we propose novel weight sharing and efficient approximation strategies. The ability to predict calibrated confidence scores not only impacts task performance but also provides a flexible means of adjusting the trade-off between precision and recall in real-world scenarios. For instance, in applications requiring high precision, predictions with lower confidence scores can be discarded. 

% \drm{This (starting with ``Our experiments\dots'') doesn't belong in this paragraph. I would give it a paragraph of it's own and place it at the end of the introduction.}.

% \ap{move all the experiment-related content to the paragraph below. The previous paragraph can focus solely on the methodology}

% \ap{instead of stating we construct 5 entity typing datasets, we could say we evaluate our approach on five specialized domains.}

% \ap{before comparing against 11B Flan-T5, we should mention our model is based on xxM T5-large model.}

% \drm{I agree with AP's suggestions here.}

% Our experiments show that filtering decoded types based on calibrated confidence scores leads to state-of-the-art performance.

We carry out extensive experiments on the UFET dataset and show that filtering decoded types based on calibrated confidence scores leads to state-of-the-art performance. Our method surpasses the previous methods in terms of both F1 score and calibration error while achieving an inference speedup of more than 50 times compared to cross-encoder methods. Furthermore, we evaluate the zero-shot and few-shot performance of our model on five specialized domains. Our model outperforms Flan-T5-XXL \cite{chung2022scaling}, an instruction-tuned large language model with 11 billion parameters in the zero-shot setting, and surpasses ChatGPT when fine-tuned on 50 examples.

%%% Local Variables:
%%% mode: latex
%%% TeX-master: "../emnlp2023"
%%% End:

\section{Related Work}

\subsection{Fine-grained Entity Typing}

\citet{ling2012fine} initiated efforts to recognize entities with labels beyond the small set of classes that is typically used in named entity recognition (NER) tasks. They proposed to formulate this task as a multi-label classification problem. More recently, \citet{choi2018ultra} extended this idea to ultra-fine entity typing and released the UFET dataset, expanding the task to include an open type vocabulary with over $10k$ classes. Interest in ultra-fine entity typing has continued to grow over the last few years. Some research efforts have focused on modeling label dependencies and type hierarchies, such as employing box embeddings \cite{onoe2021modeling} and contrastive learning \cite{zuo2022type}.

Another line of research has concentrated on data augmentation and leveraging distant supervision. For instance, \citet{dai2021ultra} obtained training data from a pretrained masked language model, while \citet{zhang2022denoising} proposed a denoising method based on an explicit noise model. \citet{li2022ultra} formulated the task as a natural language inference (NLI) problem with the hypothesis being an ``is-a'' statement. Their approach achieved state-of-the-art performance on the UFET dataset and exhibited strong generalization to unseen types, but is inefficient at inference due to the need to enumerate the entire type vocabulary.

\subsection{Probability Calibration}

Probability calibration is the task of adjusting the confidence scores of a machine learning model to better align with the true correctness likelihood. Calibration is crucial for applications that require interpretability and reliability, such as medical diagnoses. Previous research has shown that modern neural networks while achieving good task performance, are often poorly calibrated \cite{guo2017calibration, zhao2021calibrate}. One common technique for calibration in binary classification tasks is Platt scaling \cite{platt1999probabilistic}, which fits a logistic regression model on the original probabilities. \citet{guo2017calibration} proposed temperature scaling as an extension of Platt scaling in the multi-class setting. Although probability calibration has been extensively studied for single-label classification tasks \cite{jiang2020can, kadavath2022language}, it has rarely been explored in the context of fine-grained entity typing which is a multi-label classification task. To the best of our knowledge, the only exception is \citet{onoe2021modeling}, where the authors applied temperature scaling to a BERT-based model trained on the UFET dataset and demonstrated that the resulting model was reasonably well-calibrated.

%%% Local Variables:
%%% mode: latex
%%% TeX-master: t
%%% End:

\section{Methodology}

In this section, we present CASENT, a calibrated seq2seq model designed for ultra-fine entity typing. We start with the task description (\S\ref{ssec:problem_definition}) followed by an overview of the CASENT architecture (\S\ref{ssec:casent}). While the focus of this paper is on the task of entity typing, our model can be easily adapted to other multi-label classification tasks.
% begin by introducing the notation and providing a high-level overview of our model, and then discuss the training, calibration, and inference process in detail.

\subsection{Task Definition}
\label{ssec:problem_definition}

Given an entity mention $e$, we aim to predict a set of semantic types $\mathbf{t} = \{t_1, \dots, t_n\} \subset \mathcal{T}$, where $\mathcal{T}$ is a predefined type vocabulary ($|\mathcal{T}| = 10331$ for the UFET dataset). We assume each type in the vocabulary is a noun phrase that can be represented by a sequence of tokens $t = (y_1, y_2, \dots, y_{k})$. We assume the availability of a training set $\mathcal{D}_{\text{train}}$ with annotated $(e, \mathbf{t})$ pairs as well as a development set for estimating hyperparameters.

\subsection{Overview of CASENT}
\label{ssec:casent}

% \footnote{The log-probability here is normalized by target sequence length.}

\autoref{fig:model} provides an overview of our system. It consists of a seq2seq model and a calibration module. At training time, we train the seq2seq to output a ground truth type given an input entity mention by maximizing the length-normalized log-likelihood  using an autoregressive formulation 
\begin{equation}\label{eq:log_prob}
    \log p_{\theta}(t \mid e) = \frac{1}{k}\sum_{i=1}^{k} \log p_{\theta}(y_i \mid y_{<i}, e)
\end{equation}
where $\theta$ denotes the parameters of the seq2seq model.

During inference, our model takes an entity mention $e$ as input and generates a small set of candidate types autoregressively via constrained beam search by using a relatively large beam size. We then employ a calibration module to transform the raw conditional probabilities (\autoref{eq:log_prob}) associated with each candidate type into calibrated confidence scores $\hat{p}(t \mid e) \in [0, 1]$.\footnote{Here, we make a slight abuse of notation by treating $t$ as a binary random variable that indicates whether $e$ belongs to type $t$.} The candidate types whose scores surpass a global threshold are selected as the model's predictions.

The parameters of the calibration module and the threshold are estimated on the development set before each inference run (which takes place either at the end of each epoch or when the training is complete). The detailed process of estimating calibration parameters is discussed in \S\ref{subsec:calibration}.

\subsection{Training}
\label{ssec:training}

% \ap{consider changing the name of this subsection. Maybe auto-regressive model or seq2seq model?}

Our seq2seq model is trained to output a type $t$ given an input entity mention $e$. In the training set, each annotated example $(e, \mathbf{t}) \in \mathcal{D}_{\text{train}}$ with $|\mathbf{t}| = n$ ground truth types is considered as $n$ separate input-output pairs for the seq2seq model.\footnote{Note that although a training example $(e, \mathbf{t})$ is separated into $n$ input-output pairs, the forward pass at the encoder only needs to be computed once.} We initialize our model with a pretrained seq2seq language model, T5 \cite{raffel2020exploring}, and finetune it using standard maximum likelihood objective:
\begin{equation}
    \min_{\theta} \left[-\sum_{(e, \mathbf{t})\in \mathcal{D}_{\text{train}}} \sum_{t\in \mathbf{t}} \log p_{\theta} (t\mid e)\right]
\end{equation}

Our seq2seq formulation greatly simplifies the training process by eliminating the need for negative sampling, which is required by previous cross-encoder approaches \cite{li2022ultra, dai2021ultra}.

% Following previous work \cite{de2020autoregressive, de2022multilingual}

\subsection{Calibration}
\label{subsec:calibration}

At the core of our approach is a calibration module that transforms raw conditional log-probability $\log p_\theta (t\mid e)$ into calibrated confidence $\hat{p}(t\mid e)$. We will show in \autoref{sec:result} that directly applying thresholding using $p_\theta (t\mid e)$ is suboptimal as it models the distribution over target token sequences instead of the likelihood of $e$ belonging to a certain type $t$. Our approach builds on Platt scaling \cite{platt1999probabilistic} with three proposed extensions specifically tailored for the ultra-fine entity typing task: 1) incorporating model bias $p_\theta (t\mid \varnothing)$, 2) frequency-based weight sharing across types, and 3) efficient parameter estimation with sparse approximation.

\textbf{Platt Scaling}: We first consider calibration for each type $t$ separately, in which case the task reduces to a binary classification problem. A standard technique for calibrating binary classifiers is Platt scaling, which fits a logistic regression model on the original outputs. A straightforward application of Platt scaling in our seq2seq setting computes the calibrated confidence score by $\sigma(w_t \cdot \log p_\theta (t\mid e) + b_t)$, where $\sigma$ is the sigmoid function and calibration parameters $w_t$ and $b$ are estimated on the development set by minimizing the binary cross-entropy loss.

Inspired by previous work \cite{zhao2021calibrate} which measures the bias of seq2seq models by feeding them with empty inputs, we propose to learn a weighted combination of both the conditional probability $p_\theta (t\mid e)$ and model bias $p_\theta(t\mid \varnothing)$. Specifically, we propose $$\sigma\left( w^{(1)}_t \cdot\log p_\theta (t\mid e) + w^{(2)}_t \cdot \log p_\theta (t\mid \varnothing) + b_t\right)$$ as the calibrated confidence score. We will show in \autoref{sec:result} that incorporating the model bias term improves task performance and reduces calibration error.

% \begin{align}
% \begin{split}
% \hat{p}(t\mid e) = &\sigma\Big( w^{(1)}_t \cdot\log p_\theta (t\mid e)  \\
% &\qquad + w^{(2)}_t \cdot \log p_\theta (t\mid \varnothing) + b_t\Big)
% \end{split}
% \end{align}

\textbf{Multi-label Platt Scaling}: We now discuss the extension of this equation in the multi-label setting where $|\mathcal{T}| \gg 1$. A naive extension that considers each type independently would introduce $3|\mathcal{T}|$ parameters and involve training $|\mathcal{T}|$ logistic regression models on $|\mathcal{D}_{\text{dev}}|\cdot |\mathcal{T}|$ data points. To mitigate this difficulty, we propose to share calibration parameters across types based on their occurrence frequency in the dataset:
\begin{align}
\begin{split}\label{eq:final_calibration}
\hat{p}(t\mid e) &= \sigma\Big( w^{(1)}_{\phi(t)} \cdot\log p_\theta (t\mid e)  \\
&\qquad + w^{(2)}_{\phi(t)} \cdot \log p_\theta (t\mid \varnothing) + b_{\phi(t)}\Big)
\end{split}
\end{align}
where 
\begin{equation}
    \phi(t) = \big\lceil \log_2 \left( \texttt{Freq}(t) + 1\right)\big\rceil
\end{equation}
maps type $t$ to its frequency category.\footnote{On the UFET dataset, this reduces the number of calibration parameters from $30993$ to $27$.} Intuitively, rare types are more vulnerable to model bias thus should be handled differently compared to frequent types.

% Define pseudocode formatting
\renewcommand{\KwSty}[1]{\textnormal{\textcolor{blue!90!black}{\ttfamily\bfseries #1}}\unskip}
\renewcommand{\ArgSty}[1]{\textnormal{\ttfamily #1}\unskip}
\SetKwComment{Comment}{\color{green!50!black}// }{}
\renewcommand{\CommentSty}[1]{\textnormal{\ttfamily\color{green!50!black}#1}\unskip}
\newcommand{\assign}{\leftarrow}
\newcommand{\var}{\texttt}
\newcommand{\FuncCall}[2]{\texttt{\bfseries #1(#2)}}
\SetKwProg{Function}{function}{}{}
\renewcommand{\ProgSty}[1]{\texttt{\bfseries #1}}

\begin{algorithm}[t]
\small
  \caption{\small Calibration parameters estimation}
  \label{algo:calibration}
  \Function{GetCalibrationParams($\mathcal{D}_{\text{dev}}$, model, n\_groups)}{
    $\var{D} \assign \var{[[]} \KwSty{ for } \var{i} \KwSty{ in } \var{range(n\_groups)]}$\;
    \Comment{D stores the data points for estimating calibration parameters}
    \For{e, types \KwSty{in} $\mathcal{D}_{\text{dev}}$}{
        \For{t \KwSty{in} \var{model.beam\_search(e)}}{
            \var{X} $\assign$ \var{[$\log p_\theta$(t|e), $\log p_\theta$(t|$\varnothing$)]}\;
            \If{t \KwSty{in} types}{
                \var{y} $\assign$ \var{+1}
            }
            \Else{
                \var{y} $\assign$ \var{-1}
            }
            \var{D[$\phi$(t)].append((X, y))}
            
        }
    }
    \var{W} $\assign$ \var{np.zeros((n\_groups, 2))}\;
    \var{B} $\assign$ \var{np.zeros(n\_groups)}\;
    \For{i \KwSty{in} range(n\_groups)}{
        \var{W[i, :], B[i]} $\assign$ \var{FitLogisticRegression(D[i])}
    }
    \Return{W, B}\;
  }
  
\end{algorithm}
%%% Local Variables:
%%% mode: latex
%%% TeX-master: t
%%% End:

% \definecolor{color1}{HTML}{5989d4}
\definecolor{color1}{HTML}{e37c36}

% {\textcolor{color1}{[}{The explosions}\textcolor{color1}{]}\textcolor{color1}{\small event, calamity, attack, disaster} occurred on the night of October 7, against the Hilton Taba and campsites used by Israelis in Ras al-Shitan.}

% CAA reduced hRPTEC cell number and protein, induced a loss in free intracellular \entity{thiols}{chemical} and an increase in necrosis markers.

\begin{table*}[t]
    \centering
    \scalebox{0.75}{
    \begin{tabular}{llp{6.2cm}p{7.8cm}}
        \toprule
        \textbf{Dataset} & \textbf{Domain} & \textbf{Entity Types ($\mathcal{T}$)} &\textbf{Example} \\
        \midrule
        UFET & News, web articles & $10331$ types & \entity{The explosions}{event, calamity, attack, disaster} occurred on the night of October 7, against the Hilton Taba and campsites used by Israelis in Ras al-Shitan.\\
        \midrule
        WNUT2017 & Social media & \{\textcolor{color1}{{\small corporation, creative\_work, group, location, person, product}}\} & {RT @MarshmallowDoof: I did drawn the \entity{Tiger Mama}{creative\_work} @BuxbiArts}\\
        \midrule
        JNLPBA & Biomedical &  \{\textcolor{color1}{\small DNA, RNA, cell\_line, cell\_type, protein}\} & In vivo control of \textcolor{color1}{[}NF-kappa B\textcolor{color1}{]}\textcolor{color1}{\small protein} activation by I kappa B alpha. \\
        \midrule
        BC5CDR & Biomedical & \{\textcolor{color1}{\small disease, chemical}\} & In a previous phase II study with 3 - weekly bolus \entity{5-FU}{chemical}, FA and mitomycin C ( MMC ) we found a low toxicity rate and response rates comparable to those of regimens such as ELF, FAM or FAMTX, and a promising median overall survival. \\
        \midrule
        MIT-restaurant & Customer review & \{\textcolor{color1}{\small rating, amenity, location, restaurant, price, hours, dish, cuisine}\} & Can you make a reservation at \entity{pf changs}{restaurant} for tonight?\\
        \midrule
        MIT-movie & Customer review & \{\textcolor{color1}{\small actor, plot, opinion, award, year, genre, origin, director, soundtrack, relationship, character, quote}\} & An \entity{animated}{genre} movie about a criminal mastermind that attempts to steal the moon\\
        \bottomrule
    \end{tabular}}
    \caption{Dataset statistics and examples. Only UFET has multiple types for each entity mention.}
    \label{tab:dataset}
\end{table*}

Furthermore, instead of training logistic regression models on all $|\mathcal{D}_{\text{dev}}|\cdot |\mathcal{T}|$ data points, we propose a sparse approximation strategy that only leverages candidate types generated by the seq2seq model via beam search.\footnote{This reduces the maximum number of calibration data points to $|\mathcal{D}_{\text{dev}}|\times \text{BeamSize}$.} This ensures that the entire calibration process retains the same time complexity as a regular evaluation run on the development set. The pseudo code for estimating calibration parameters is outlined in \autoref{algo:calibration}. Once the calibration parameters have been estimated, we select the optimal threshold by running a simple linear search.

\subsection{Inference}
\label{ssec:inference}

At test time, given an entity mention $e$, we employ constrained beam search to generate a set of candidate types autoregressively. Following previous work \cite{de2020autoregressive, de2022multilingual}, we pre-compute a prefix trie based on $\mathcal{T}$ and force the model to select valid tokens during each decoding step. Next, we compute the calibrated confidence scores using \autoref{eq:final_calibration} and discard types whose scores fall below the threshold.

In \autoref{sec:result}, we also conduct experiments on single-label entity typing tasks. In such cases, we directly score each valid type using \autoref{eq:final_calibration} and select the type with the highest confidence score.

%%% Local Variables:
%%% mode: latex
%%% TeX-master: "../emnlp2023"
%%% End:

\section{Experiments} \label{sec:result}

\subsection{Datasets} \label{subsec:dataset}

We use the UFET dataset \cite{choi2018ultra}, a standard benchmark for ultra-fine entity typing. This dataset contains $10331$ entity types and is curated by sampling sentences from GigaWord \cite{parker2011english}, OntoNotes \cite{hovy2006ontonotes} and web articles \cite{singh2012wikilinks}.

To test the out-of-domain generalization abilities of our model, we construct five entity typing datasets for three specialized domains. We derive these from existing NER datasets, WNUT2017 \cite{derczynski2017results}, JNLPBA \cite{collier2004introduction}, BC5CDR \cite{wei2016assessing}, MIT-restaurant and MIT-movie.\footnote{\url{https://groups.csail.mit.edu/sls/downloads/}} We treat each annotated entity mention span as an input to our entity typing model. WNUT2017 contains user-generated text from platforms such as Twitter and Reddit. JNLPBA and BC5CDR are both sourced from scientific papers from the biomedical field. MIT-restaurant and MIT-movie are customer review datasets from the restaurant and movie domains respectively. \autoref{tab:dataset} provides the statistics and an example from each dataset.
% Furthermore, we also evaluate the out-of-domain generalization performance of our model in zero-shot and few-shot settings.

\subsection{Implementation}

We initialize the seq2seq model with pretrained T5-large \cite{raffel2020exploring} and finetune it on the UFET training set with a batch size of $8$. We optimize the model using Adafactor \cite{shazeer2018adafactor} with a learning rate of 1e-5 and a constant learning rate schedule. The constrained beam search during calibration and inference uses a beam size of $24$. We mark the entity mention span with a special token and format the input according to the template \texttt{``\{CONTEXT\} </s> \{ENTITY\} is </s>''}. Input and the target entity type are tokenized using the standard T5 tokenizer.

\subsection{Baselines}

We compare our method to previous state-of-the-art approaches, including multi-label classifier-based methods such as BiLSTM \cite{choi2018ultra}, BERT, Box4Types \cite{onoe2021modeling} and MLMET \cite{dai2021ultra}. In addition, we include a bi-encoder model, UniST \cite{huang2022unified} as well as the current state-of-the-art method, LITE \cite{li2022ultra}, which is based on a cross-encoder architecture.

We also compare with ChatGPT \footnote{We use the \texttt{gpt-3.5-turbo-0301} model available via the OpenAI API.} and Flan-T5-XXL \cite{chung2022scaling}, two large language models that have demonstrated impressive few-shot and zero-shot performance across various tasks. For the UFET dataset, we randomly select a small set of examples from the training set as demonstrations for each test instance. Instruction is provided before the demonstration examples to facilitate zero-shot evaluation. Furthermore, for the five cross-domain entity typing datasets, we supply ChatGPT and Flan-T5-XXL with the complete list of valid types. Sample prompts are shown in \autoref{app:prompt}.

% \ap{talk about the evaluation metric}
\begin{table}[]
    \centering
    \small
    \begin{tabular}{lccc}
        \toprule
         Method & {P} & {R} & {F1} \\
        \midrule
        \textit{Few-shot methods} \\
        \midrule
        ChatGPT (0-shot) & 55.5 & 10.5 & 17.6 \\
        ChatGPT (8-shot) & 46.7 & 34.9 & 40.0 \\
        ChatGPT (16-shot) & 47.8 & 36.7 & 41.5 \\
        ChatGPT (32-shot) & 45.9 & 37.3 & 41.2 \\
        \midrule
        \textit{Supervised methods} \\
        \midrule
        BiLSTM \cite{choi2018ultra} & 47.1 & 24.2 & 32.0 \\
        BERT \cite{onoe2019learning} &  51.6 & 33.0 & 40.2 \\
        Box4Types \cite{onoe2021modeling} & 52.8 & 38.8 & 44.8 \\
        MLMET \cite{dai2021ultra} & 53.6 & 45.3 & 49.1 \\
        UniST \cite{huang2022unified} & 50.2 & 49.6 & 49.9 \\
        LITE \cite{li2022ultra} & 52.4 & 48.9 & \underline{50.6} \\
        % \midrule
        % CASENT$_{base}$ (w/o calibration) & 45.4 & 45.3 & 45.4 \\
        % CASENT (w/o calibration) & 49.9 & 45.0 & 47.3 \\
        % CASENT$_{\text{base}}$ (Ours) & 50.4 & 48.8 & 49.6 \\
        CASENT (Ours) & 53.3 & 49.5 & \textbf{51.3} \\
        \bottomrule
    \end{tabular}
    \caption{Macro-averaged precision, recall and F1 score (\%) on the UFET test set. The model with highest F1 score is shown in \textbf{bold} and the second best is \underline{underlined}. }
    \label{tab:sota}
\end{table}

\begin{table*}[t]
    \centering
    \scalebox{0.78}{
    \begin{tabular}{lcccccc}
        \toprule
        \multirow{2}{*}{Method}&
        {Social Media} & \multicolumn{2}{c}{Biomedical} & \multicolumn{2}{c}{Customer Review} \\
        \cmidrule(lr){2-2} \cmidrule(lr){3-4} \cmidrule(lr){5-6}
        & WNUT 2017 & JNLPBA & BC5CDR & MIT-restaurant & MIT-movie & Avg.  \\
\midrule
\textit{Zero-shot methods} \\
\midrule
Random & 16.7 & 20.0 & 50.0 & 12.5 & 8.3 & 21.5 \\
Flan-T5-XXL & 62.9 & 71.8 & 63.0 & 39.6 & 45.4 & 56.5\\
ChatGPT                                      & \underline{76.3} & \underline{85.4} & 96.7  & \underline{80.3} & \underline{77.2} & \underline{83.2} \\
LITE (Li et al. 2022)      
& 67.0  & 74.9 & 96.1 & 47.7 & 54.5 & 68.0 \\
CASENT (no finetuning, no calibration)     & 65.5 &  79.2 & \underline{98.2}& 52.9  & 51.2 & 69.4 \\
\midrule
\textit{Few-shot methods} \\
\midrule
RoBERTa-large (finetuned on 50 examples)            & 65.5 & 85.1  & 96.2& 75.0 & 69.9 & 78.3 \\
CASENT (no finetuning, calibration on dev) & 74.2 & 84.0 & \underline{98.2}  & 68.5 & 71.7 & 79.3 \\
CASENT (finetuned on 50 examples)           & \textbf{77.3}& \textbf{92.2} & \textbf{98.8}  & \textbf{81.8} & \textbf{86.2} & \textbf{87.2} \\
        \bottomrule
    \end{tabular}}
    \caption{Test set accuracy on five specialized domain entity typing datasets derived from existing NER datasets. The best score is shown in \textbf{bold} and the second best is \underline{underlined}. The results of LITE are obtained by running inference using the model checkpoint provided by the authors. }
    \label{tab:ner_typing}
\end{table*}

\section{Results}
\label{sec:results}

\subsection{UFET}

In \autoref{tab:sota}, we compare our approach with a suite of baselines and state-of-the-art systems on the UFET dataset. Our approach outperforms LITE \cite{li2022ultra}, the current leading system based on a cross-encoder architecture, with a $0.7\%$ improvement in the F1 score. Among the fully-supervised models, cross-encoder models demonstrate superior performance over both bi-encoder methods and multi-label classifier-based models.

ChatGPT exhibits poor zero-shot performance with significantly low recall. However, it is able to achieve comparable performance to a BERT-based classifier with a mere $8$ few-shot examples. Despite this, its performance still lags behind recent fully supervised models. 

\subsection{Out-of-domain Generalization}

We evaluate the out-of-domain generalization performance of different models on the five datasets discussed in \S\ref{subsec:dataset}. The results are presented in \autoref{tab:ner_typing}. It is important to note that we don't compare with multi-label classifier models like Box4Types and MLMET that treat types as integer indices, as they are unable to generalize to unseen types. 

In the zero-shot setting, LITE and CASENT are trained on the UFET dataset and directly evaluated on the target test set. Flan-T5-XXL and ChatGPT are evaluated by formulating the task as a classification problem with all valid types as candidates. As shown in \autoref{tab:ner_typing}, ChatGPT demonstrates superior performance with a large margin compared to other models. This highlights ChatGPT's capabilities on classification tasks with a small label space. Our approach achieves comparable results to LITE and significantly outperforms Flan-T5-XXL, despite having less than $10\%$ of its parameters.

We also conduct experiments in the few-shot setting, where either a small training set or development set is available.  We first explore re-estimating the calibration parameters of CASENT on the target development set by following the process discussed in \S\ref{subsec:calibration} without weight sharing and sparse approximation.\footnote{The number of calibration parameters is $3|\mathcal{T}|$, which is less than $40$ on all five datasets.} Remarkably, this re-calibration process, without any finetuning, results in an absolute improvement of $+9.9\%$ and comparable performance with ChatGPT on three out of five datasets. When finetuned on $50$ randomly sampled examples, our approach outperforms ChatGPT and a finetuned RoBERTa model by a significant margin, highlighting the benefits of transfer learning from the ultra-fine entity typing task.

% In the zero-shot setting, both CASENT and LITE achieves reasonable performance 

%%% Local Variables:
%%% mode: latex
%%% TeX-master: "../emnlp2023"
%%% End:

\begin{table*}[h]
    \centering
    % \small
    \scalebox{0.77}{\begin{tabular}{llccccc}
        \toprule
        Method & Calibration Method & {Test-F1} (\%) & {Test-ECE} (\%)   & {Test-TCE} (\%) & {Dev-TCE} (\%) \\
        \midrule
        Box4Types & Temperature scaling & 44.8 & - & - & 11.19 \\
        LITE   & - & 50.6 & 52.36 & 52.36 & 52.56\\
        % \midrule
        % CASENT$_{base}$ (w/o calibration) & 45.4 & 45.3 & 45.4 \\
        % CASENT (w/o calibration) & 49.9 & 45.0 & 47.3 \\
        % CASENT$_{\text{base}}$ (Ours) & 50.4 & 48.8 & 49.6 \\
        \midrule
          CASENT & Eq. \ref{eq:final_calibration}  & \textbf{51.3} & \textbf{1.23}   &   \textbf{9.75} & \textbf{9.38} \\
& Eq. \ref{eq:final_calibration} without the model bias term $p_\theta(t\mid \varnothing)$ & 49.4 & 3.87   &  20.34 & 14.76 \\
& Eq. \ref{eq:final_calibration} with $\phi(t) = t$ (independent weights) & 48.8 & 7.37   &   57.00 & 9.72 \\
& Eq. \ref{eq:final_calibration} with $\phi(t) = t_0$ (all types share same weights)  & 47.8 & 3.89   &  34.57 & 36.29 \\
& $p_\theta(t\mid e)$ (no calibration)  & 47.3 & 12.19  & 118.16 & 100.31 \\
        \bottomrule 
    \end{tabular}}
    \caption{Macro F1, ECE (Expected Calibration Error) and TCE (Total Calibration Error) on the UFET dataset. ECE and TCE are computed using 10 bins. The best score is shown in \textbf{bold}. \citet{onoe2021modeling} only reported calibration results on the dev set thus the results of Box4Types on the test set are not included.}
    \label{tab:calibration}
\end{table*}

\section{Analysis}

\subsection{Calibration}

\begin{figure}
    \centering
    \includegraphics[width=\linewidth]{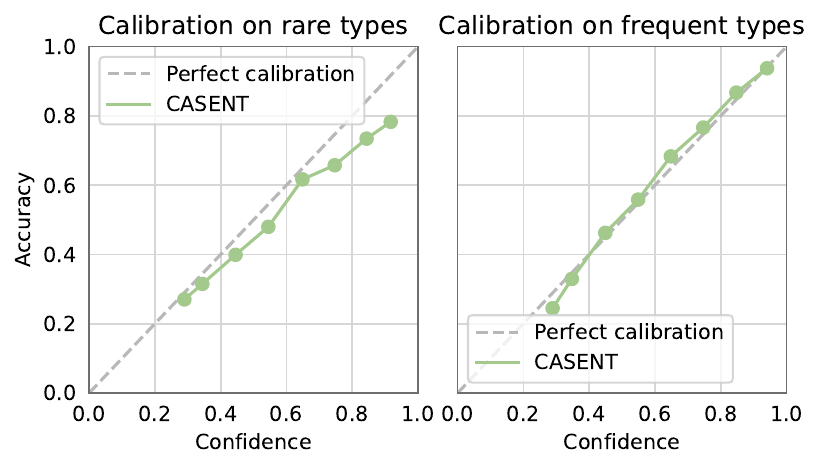}
    \caption{Reliability diagrams of CASENT on the UFET test set. The left diagram represents rare types with fewer than 10 occurrences while the right diagram represents frequent types.}
    \label{fig:calibration_curve}
\end{figure}

% \ap{add a citation for ECE and TCE}
\autoref{tab:calibration} presents the calibration error of different approaches. We report Expected Calibration Error (ECE) and Total Calibration Error (TCE) which measures the deviation of predicted confidence scores from empirical accuracy. Interestingly, we observe that the entailment scores produced  by LITE, the state-of-the-art cross-encoder model, are poorly calibrated. Our approach achieves slightly lower calibration error than Box4Types, which applies temperature scaling \cite{guo2017calibration} to the output of a BERT-based classifier. \autoref{fig:calibration_curve} displays the reliability diagrams of CASENT for both rare types and frequent types. As illustrated by the curve in the left figure, high-confidence predictions for rare types are less well-calibrated.

\subsection{Ablation Study}

We also perform an ablation study to investigate the impacts of various design choices in our proposed calibration method. \autoref{tab:calibration} displays the results of different variants of CASENT. A vanilla seq2seq model without any calibration yields both low task performance and high calibration error, highlighting the importance of calibration. Notably, a naive extension of Platt scaling that considers each type independently leads to significant overfitting, illustrated by an absolute difference of $47.28\%$ TCE between the development and test sets. Removing the model bias term also has a negative impact on both task performance and calibration error.

\subsection{Choice of Seq2seq Model}

\begin{table}[]
    \centering
    \small
    \begin{tabular}{llc}
        \toprule
         Method & \# params &  F1 \\
        \midrule
        T5-small    & \multirow{2}{*}{80M}      & 40.9 \\
        T5-small + CASENT & &  47.2 \\
        \midrule
        T5-base          & \multirow{2}{*}{250M} & 45.4 \\
        T5-base + CASENT & & 49.6 \\
        \midrule
        T5-large     & \multirow{2}{*}{780M}    &  47.3 \\
        T5-large + CASENT &    &      51.3    \\
        \midrule
        T5-3B           & \multirow{2}{*}{3B} & 48.6    \\
        T5-3B + CASENT  &  &  51.4\\
        \bottomrule
    \end{tabular}
    \caption{Macro F1 score (\%) of CASENT on the UFET test set with different T5 variants. }
    \label{tab:base_model}
\end{table}

In \autoref{tab:base_model}, we demonstrate the impact of calibration on various T5 variants. Our proposed calibration method consistently brings improvement across models ranging from 80M parameters to 3B parameters. The most substantial improvement is achieved with the smallest T5 model.

\subsection{Training and Inference Efficiency}

\begin{table}[]
    \centering
    \scalebox{0.8}{\begin{tabular}{@{}lccc@{}}
        \toprule
         Method & Training Time &  Inference Latency & GPU Mem. \\
        \midrule
        MLMET & $18$0h$^\dag$ & $0.02\pm 0.05 $s & $0.5$Gb \\
        LITE & $40$h$^\dag$ & $23.1 \pm 5.73$s & $1.4$Gb \\
        CASENT & $6$h & $0.39 \pm 0.04$s & $2.8$Gb \\
        \bottomrule
    \end{tabular}}
    \caption{Training time, inference latency and inference time GPU memory usage estimated on a single NVIDIA RTX A6000 GPU. Inference time statistics are estimated using 100 random UFET examples. Results marked by $\dag$ are reported by \citet{li2022ultra}.}
    \label{tab:efficiency}
\end{table}

In \autoref{tab:efficiency}, we compare the efficiency of our method with previous state-of-the-art systems. Remarkably, CASENT only takes 6 hours to train on a single GPU, while previous methods require more than 40 hours. While CASENT achieves an inference speedup of over 50 times over LITE, it is still considerably slower than MLMET, a BERT-based classifier model. This can be attributed to the need for autoregressive decoding in CASENT.

\subsection{Impact of Beam Size}

\begin{figure}
    \centering
    \includegraphics[width=\linewidth]{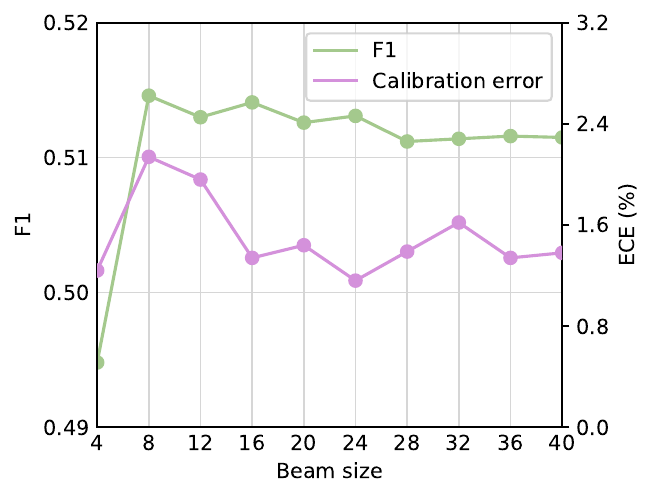}
    \caption{Test set Macro F1 score and Expected Calibration Error (ECE) with respect to the beam size on the UFET dataset.}
    \label{fig:beam_size}
\end{figure}

Given that the inference process of CASENT relies on constrained beam search, we also investigate the impact of beam size on task performance and calibration error. As shown in \autoref{fig:beam_size}, a beam size of 4 results in a low calibration error but also low F1 scores, as it limits the maximum number of predictions. CASENT consistently maintains high F1 scores with minor fluctuations for beam sizes ranging from 8 to 40. On the other hand, a beam size between 8 and 12 leads to high calibration errors. This can be attributed to our calibration parameter estimation process in \autoref{algo:calibration}, which approximates the full $|\mathcal{D}_{\text{dev}}|\cdot|\mathcal{T}|$ calibration data points using model predictions generated by beam search. A smaller beam size leads to a smaller number of calibration data points, resulting in a suboptimal estimation of calibration parameters.

% \subsection{Case Study}
% \yanlinf{TODO}

%%% Local Variables:
%%% mode: latex
%%% TeX-master: "../emnlp2023"
%%% End:

\section{Conclusion}

%\drm{This conclusion is fine, but doesn't provide a broader vision. If we are short on space, consider leaving it out.}

% We present CASENT, a calibrated seq2seq model that predict ultra-fine semantic types with calibrated confidence scores. Our model is trained with standard maximum likelihood criteria and performs inference via constrained beam search with a novel probability calibration method. This proposed approach outperforms previous methods on in the ultra-fine entity typing task and exhibits strong generalization performance to unseen domains. 

% \drm{Here's my version}

Engineering decisions often involve a tradeoff between efficiency and accuracy. CASENT simultaneously improves upon the state-of-the-art in both dimensions while also being conceptually elegant. The heart of this innovation is a constrained beam search with a novel probability calibration method designed for seq2seq models in the multi-label classification setting. Not only does this method outperform previous methods---including ChatGPT and the existing fully-supervised methods---on ultra-fine entity typing, but it also exhibits strong generalization capabilities to unseen domains.

% \ap{Would it be possible to include paper URLs in the references?}

%%% Local Variables:
%%% mode: latex
%%% TeX-master: "../emnlp2023"
%%% End:

\section{Limitations}

While our proposed CASENT model shows promising results on ultra-fine entity typing tasks, it does have certain limitations. Our experiments were conducted using English language data exclusively and it remains unclear how well our model would perform on data from other languages. In addition, our model is trained on the UFET dataset, which only includes entity mentions that are identified as noun phrases by a constituency parser. Consequently, certain types of entity mentions such as song titles are excluded. The performance and applicability of our model might be affected when dealing with such types of entity mentions. Future work is needed to adapt and evaluate the proposed approach in other languages and broader scenarios.

\section*{Acknowledgements}

This material is based on research sponsored by the Air Force Research
Laboratory under agreement number FA8750-19-2-0200. The U.S. Government
is authorized to reproduce and distribute reprints for Governmental
purposes notwithstanding any copyright notation thereon. The views and
conclusions contained herein are those of the authors and should not be
interpreted as necessarily representing the official policies or
endorsements, either expressed or implied, of the Air Force Research
Laboratory or the U.S. Government.

\bibliographystyle{acl_natbib}
\bibliography{bib/anthology, bib/custom}

\clearpage
\appendix
\section{ChatGPT / Flan-T5 Prompts} \label{app:prompt}

% \begin{adjustbox}{width=\linewidth}
% \begin{tcolorbox}[colback=white, colframe=black, rounded corners]

Below is a sample prompt for ChatGPT and Flan-T5-XXL for the five out-of-domain datasets: 
\begin{tcolorbox}
\texttt{\footnotesize  \fontfamily{lmss}\selectfont Instruction: Identify the type of the entity mention tagged by <mark>. Output the type directly and do not write any explanation.\\
Choices: DNA, RNA, cell\_line, cell\_type, protein\\
Entity: Number of <mark>glucocorticoid receptors</mark> in lymphocytes and their sensitivity to hormone action . \\
Label:}
\end{tcolorbox}
% \end{adjustbox}

For the UFET dataset, it is not feasible to provide the model with the entire type vocabulary. Instead we provides demonstration examples sampled from the training set. Below is a sample prompt with two demonstration examples:
\begin{tcolorbox}
\texttt{\footnotesize  Instruction: Predict the fine-grained entity types for the entity mention tagged by <mark>. Separate the types with commas. \\\\
Entity: <mark>He</mark> get 's zero from Arafat , '' said Benjamin Begin , the science minister .\\
Labels: academician, scientist, person \\\\
Entity: President Obama 's surprise proposal to cancel the \$ 108 billion moon program and the jobs that go with <mark>it</mark> triggered an uproar in Texas , Florida and other states with space - related industries .\\
Labels: work, job, bill\\\\
Entity: On <mark>late Monday night</mark> , 30th Nov 2009 , Bangladesh Police arrested Rajkhowa somewhere near Dhaka .\\
Labels:}
\end{tcolorbox}

\end{document}